%% file: main.tex
\documentclass{article}
\ifdefined\pdfminorversion\pdfminorversion=7\fi
\usepackage[utf8]{inputenc}
\usepackage[T1]{fontenc}
\usepackage{amsmath}
\usepackage{newtxtext,newtxmath}
\PassOptionsToPackage{numbers,sort&compress}{natbib}
\usepackage[preprint]{neurips_2026}

\usepackage{amssymb}
\usepackage{graphicx}
\usepackage{booktabs,multirow,makecell,array}
\usepackage{fix-cm}
\usepackage[table]{xcolor}
\usepackage{adjustbox}
\usepackage{caption}
\usepackage{wrapfig}
\usepackage{placeins}
\usepackage{needspace}
\usepackage{microtype}
\usepackage{hyperref}
\usepackage{url}
\definecolor{baselinecolor}{gray}{.9}

\hypersetup{
  pdftitle={Adaptive Visual KV Cache Reuse for Efficient Multi-Turn Iterative Multimodal Reasoning},
  pdfauthor={Mengzhao Wang, Yanli Ji, Wangmeng Zuo, Chongjun Tu},
  colorlinks=false,
  pdfborder={0 0 1},
  citebordercolor={0 1 0}
}

\title{Adaptive Visual KV Cache Reuse for Efficient\\
Multi-Turn Iterative Multimodal Reasoning}
\author{
Mengzhao Wang$^{1}$, Yanli Ji$^{1,2\dagger}$, Wangmeng Zuo$^{3}$, Chongjun Tu$^{4}$ \\
$^{1}$Sun Yat-sen University \quad
$^{2}$Shenzhen Loop Area Institute \\
$^{3}$Harbin Institute of Technology (HIT) \quad
$^{4}$Fudan University \\
}
\makeatletter
\renewcommand{\@noticestring}{}
\renewcommand{\maketitle}{%
  \begingroup
  \renewcommand{\thefootnote}{\fnsymbol{footnote}}%
  \thispagestyle{plain}%
  \begin{center}
    \vspace*{-0.6em}
    {\LARGE\bfseries \@title\par}
    \vspace{0.35em}
    {\normalsize \@author\par}
    \vspace{0.5em}
  \end{center}
  \endgroup
}
\makeatother

\newcommand{\PKR}{\textsc{PKR}}
\newcommand{\IVE}{\textsc{IVE}}
\newcommand{\clip}{\operatorname{clip}}
\newcommand{\ours}[1]{\cellcolor{baselinecolor}{#1}}
\newcolumntype{L}[1]{>{\raggedright\arraybackslash}p{\dimexpr#1\textwidth-2\tabcolsep\relax}}
\newcolumntype{C}[1]{>{\centering\arraybackslash}p{\dimexpr#1\textwidth-2\tabcolsep\relax}}

\begin{document}
\maketitle
\begingroup
\renewcommand{\thefootnote}{\fnsymbol{footnote}}
\footnotetext[2]{Corresponding author.}
\endgroup
\suppressfloats[t]

\begin{abstract}
Multi-turn visual reasoning acquires evidence through repeated cropping and local
inspection, but overlapping observations lead to redundant visual encoding and
prefill. We propose an adaptive visual KV cache reuse framework that reduces
repeated processing while retaining access to additional image detail. The central
difficulty is that skipping visual prefill also removes the interaction between a
new observation and the current reasoning: historical keys remain bound to old
positions, while historical values have not incorporated the intent formed between
crops. Our framework first selects full re-encoding or regional reuse according to
crop size and cache availability. On the reuse path, Positional Key Rebinding
(\PKR) corrects attention matching at the new positions, and Intent-guided Value
Enhancement (\IVE) uses pre-crop reasoning attention to strengthen content relevant
to the current objective. Source states from fully re-encoded crops are retained
for subsequent retrieval and reuse. The framework requires no additional training;
on reuse hits, it bypasses the visual encoder and prefill for the selected visual
tokens, then resumes normal text reasoning. Experiments show that our method reduces
inference time while preserving the overall accuracy of multi-turn visual reasoning
that relies on intermediate crop images.
\end{abstract}

\section{Introduction}
\label{sec:introduction}

Multimodal models increasingly acquire visual evidence during reasoning rather
than interpret images once, using existing observations to locate, crop, and
magnify regions for further inspection \citep{lai2026minio3}.
V* uses LLM-guided visual search to locate fine-grained evidence, Visual Sketchpad integrates visual tools
and their outputs, and ICoT interleaves image patches with textual rationales
\citep{wu2024vstar,hu2024sketchpad,gao2025icot}.
DeepEyes and Pixel Reasoner learn active inspection through reinforcement
learning, while Mini-o3 extends visual search to longer, more diverse trajectories
\citep{zheng2026deepeyes,wang2025pixelreasoner,lai2026minio3}.
Together, these advances make perception an ongoing process of acquiring
evidence as reasoning progresses.

\begin{figure}[t]
  \centering
  \includegraphics[width=0.96\linewidth]{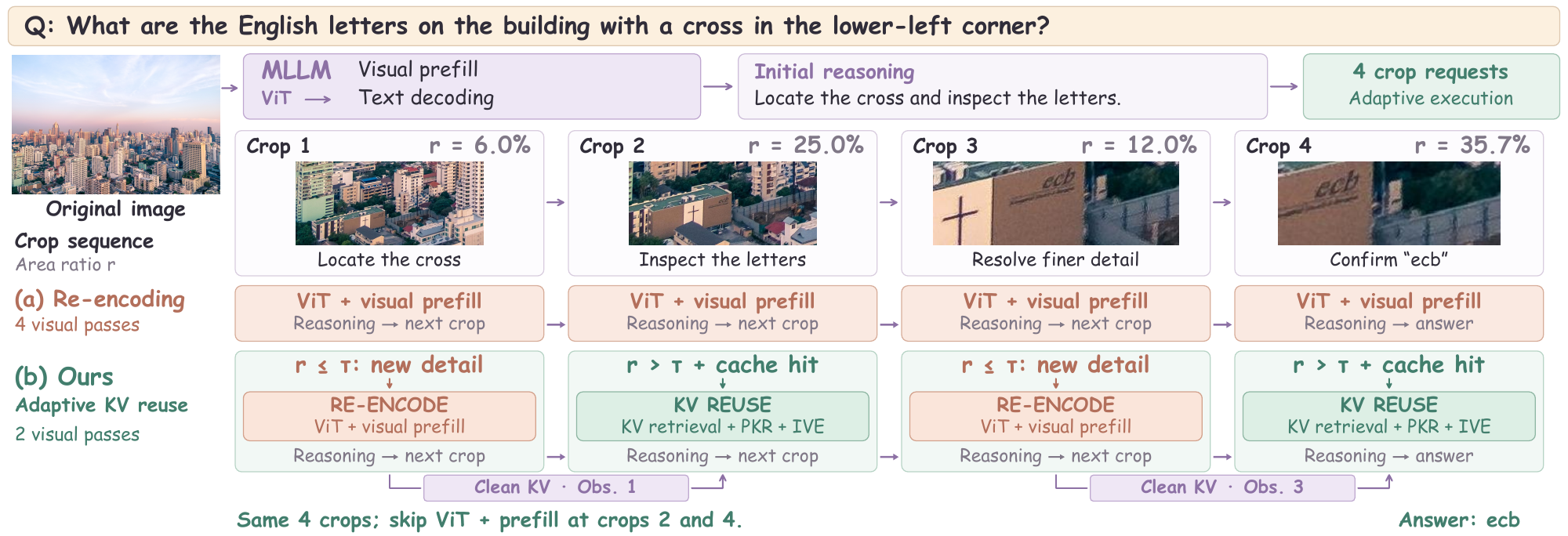}
  \caption{\textbf{Repeated re-encoding versus adaptive visual KV cache reuse.}
  For the same four crop requests, (a) full re-encoding repeats visual encoding and
  prefill at every step, whereas (b) adaptive routing re-encodes crops 1 and 3 for
  new detail and reuses their cached states for crops 2 and 4 through \PKR{} and
  \IVE{}, skipping visual encoding and prefill for two crops.}
  \label{fig:introduction}
  \vspace{-1.2em}
\end{figure}

Advances in model design and optimization improve performance
\citep{ye2026betadartspp,ye2025stimulativepp,ye2025mow}, but multi-turn visual
reasoning also demands efficient inference. Standard pipelines encode each
returned crop and prefill its visual tokens (Figure~\ref{fig:introduction}(a)),
repeatedly processing overlapping regions without using their existing states.
Visual token compression reduces computation within an observation
\citep{chen2024fastv,zhang2025sparsevlm,yang2025visionzip,shang2025prumerge},
whereas multi-turn reasoning also permits KV reuse across observations.
Since cropping and magnifying original pixels remains essential for new detail,
we study how to reuse prior visual computation across successive crops while
retaining full re-encoding for further inspection.

Such reuse requires historical visual states to fit the current reasoning context.
Standard visual prefill constructs keys at their new insertion positions and lets
visual representations interact with the preceding reasoning through successive
layers. Direct reuse omits both computations (Figure~\ref{fig:motivation}). Historical keys retain
rotary position embedding (RoPE) encodings for earlier positions, changing query--key matching
and potentially disturbing attention allocation between visual evidence and
reasoning text \citep{su2021roformer}. Even after positions are corrected, the
values still come from an earlier observation: intervening reasoning may have
shifted the objective from locating an object to reading text or identifying an
attribute, but this new crop intent did not participate in constructing the cached
values. Visual KV reuse must thus address positional mismatch and evolving crop
intent so that existing visual content can support the next reasoning step.

Based on this analysis, we propose \emph{adaptive visual KV cache reuse}
(Figure~\ref{fig:introduction}(b)). The framework uses relative crop size and cache
availability to choose between image re-encoding and regional KV retrieval.
For reused regions, \emph{Positional Key Rebinding} (\PKR) applies rotary encoding
to clean source keys at the current insertion positions, restoring their intended
relative positions for attention. \emph{Intent-guided Value Enhancement} (\IVE) uses pre-crop reasoning
attention to the parent observation to strengthen values relevant to the current
objective. Together, they connect historical visual states to the new reasoning
step, bypassing visual encoding and prefill for the selected tokens. The framework
requires no additional training and leaves the model's crop-decision mechanism
unchanged, although realized reasoning and crop trajectories may differ.

Our main contributions are as follows:
\begin{itemize}
    \item \textbf{Adaptive visual KV cache reuse.}
    We select between regional reuse and full re-encoding based on crop size
    and cache availability, reducing repeated computation while retaining
    access to new visual detail.

    \item \textbf{Positional Key Rebinding.}
    We analyze positional mismatch in direct reuse and rebind cached keys
    to their current insertion positions, restoring the intended relative
    positions in attention.

    \item \textbf{Intent-guided Value Enhancement.}
    We use reasoning between crops to strengthen values relevant to the
    current objective, without rerunning prefill for the selected visual
    tokens.

    \item \textbf{Accuracy and efficiency.}
    Across five benchmarks, our method preserves or improves overall
    accuracy while achieving $1.30$--$1.40\times$ end-to-end speedups.
\end{itemize}

\section{Related Work}
\label{sec:related}

\subsection{Iterative Visual Reasoning}

Iterative visual reasoning lets models inspect an image further based on
intermediate reasoning. V* uses language-guided search to locate fine-grained
evidence in high-resolution images, Visual Sketchpad incorporates visual tool calls
and their outputs, and ICoT organizes reasoning through interleaved image patches
and text \citep{wu2024vstar,hu2024sketchpad,gao2025icot}.
DeepEyes and Pixel Reasoner learn visual tool use through reinforcement learning,
while Mini-o3 expands the reasoning patterns and interaction turns used in visual
search \citep{zheng2026deepeyes,wang2025pixelreasoner,lai2026minio3}.
These methods advance decisions about when and where to look. We focus on the
computation after a crop request: retaining the existing observation policy,
reusing previously computed regional states, and adapting their use to the current
reasoning objective.

\subsection{Efficient Inference and KV Cache Reuse}

FastV, SparseVLM, VisionZip, and LLaVA-PruMerge reduce visual-token computation
through pruning or aggregation
\citep{chen2024fastv,zhang2025sparsevlm,yang2025visionzip,shang2025prumerge},
while VASparse combines visual sparsification with hallucination mitigation
\citep{zhuang2025vasparse}. H$_2$O, StreamingLLM, and SnapKV retain cache states
based on attention, attention sinks, and prompt relevance, respectively
\citep{zhang2023h2o,xiao2024streamingllm,li2024snapkv}.
For changing contexts, CacheBlend selectively recomputes cached text chunks
\citep{yao2024cacheblend}; MPIC combines position-independent multimodal caching
with selective recomputation \citep{zhao2025mpic}; Kamera adds low-rank contextual
compensation to positional relocation \citep{ma2026kamera}; and CONDUIT uses
attention and values to select visual states for refresh \citep{chen2026conduit}.
We focus on regional visual KV reuse across successive crops, adapting cached
states through adaptive routing, key-position rebinding, and intent-guided value
enhancement while retaining full re-encoding when new detail is needed.

\input{sections/observations_motivation}

\section{Method}
\label{sec:method}

\begin{figure}[t]
  \centering
  \includegraphics[width=\linewidth]{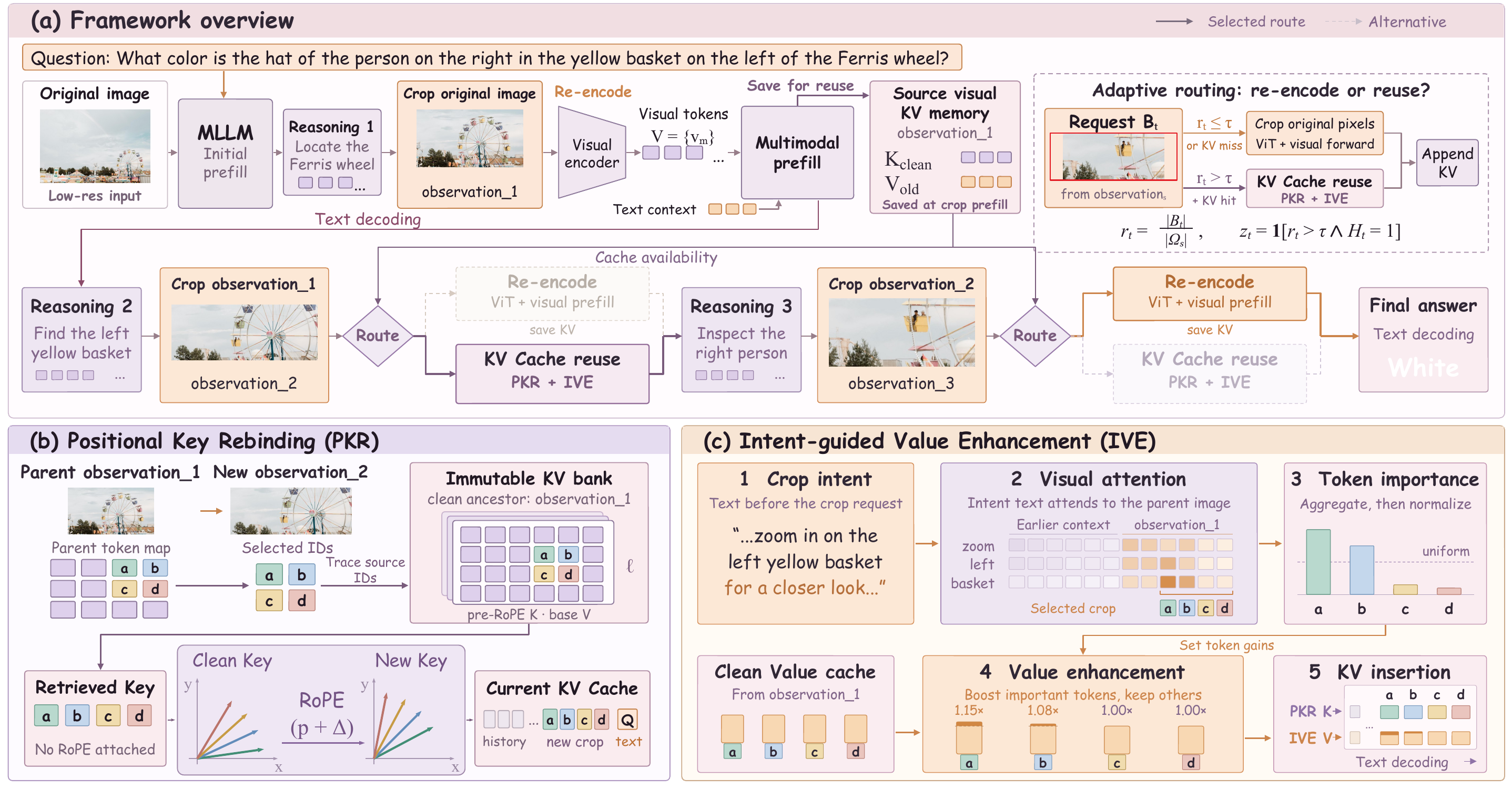}
  \caption{\textbf{Adaptive visual KV cache reuse.}
  (a) Requests are routed to full re-encoding or regional cache reuse.
  Fully re-encoded crops populate an immutable source bank.
  (b) \PKR{} traces selected tokens to clean source keys and applies RoPE at
  the target positions $\mathbf p_j^{\mathrm{new}}$.
  (c) \IVE{} uses pre-crop reasoning attention to derive token-wise gains and
  enhance base values before insertion. Subsequent text decoding proceeds normally.}
  \label{fig:framework}
      \vspace{-1.2em}
\end{figure}

Our method handles each crop request by deciding whether new visual information
is needed from the original image, then supplying the corresponding visual KV
states for subsequent reasoning. As shown in Figure~\ref{fig:framework}(a), the
framework has two paths. \textbf{Re-encoding} crops the target region from
the original image and performs visual encoding and prefill.
\textbf{Cache reuse} retrieves existing visual states for the region, adapts
them through Positional Key Rebinding (\PKR) and Intent-guided Value Enhancement
(\IVE), and resumes text generation.

\paragraph{Adaptive routing.}
At crop step $t$, the model requests region $B_t\subseteq\Omega_s$ from
observation $O_s$ ($s<t$), which we call
the parent observation. Routing depends on the crop's area relative to its
parent (Figure~\ref{fig:framework}(a)). Smaller crops typically require further magnification, so we crop and
re-encode original-image pixels to obtain finer visual detail. Larger crops
preferentially reuse available states to reduce repeated processing of
overlapping content. Let $\Omega_s$ denote the parent's valid image region and
$H_t\in\{0,1\}$ indicate whether the required visual cache entries and regional
mappings are available. Both areas are measured in the same parent-image
coordinates, with $|\Omega_s|>0$. We select the path as
\begin{equation}
  r_t=\frac{|B_t|}{|\Omega_s|},\qquad
  z_t=\mathbf{1}\!\left[r_t>\tau\ \land\ H_t=1\right].
  \label{eq:routing}
\end{equation}
Here $|\cdot|$ denotes area, $\mathbf1[\cdot]$ is the indicator function,
and $\tau$ is the area-ratio threshold; $z_t=1$ selects cache reuse. When $r_t\leq\tau$ or the required cache is
unavailable, we use full re-encoding: map the target region to the original image,
crop and magnify its pixels, and perform visual encoding and visual-token prefill.

\paragraph{Saving and retrieving visual KV caches.}
As shown in Figure~\ref{fig:framework}(a,b), full re-encoding stores clean
pre-RoPE keys, unenhanced base values, and regional mappings at every layer.
For each crop, we trace these mappings to its fully prefilled source, including
an ancestor if the parent was itself obtained through reuse. Let $j=1,\ldots,N$
index the selected visual tokens. We write $\bar{\mathbf k}_j$ and
$\mathbf v_j^0$ for the retrieved entries, already aligned to crop token $j$
through $\psi(j)$. The source bank remains unchanged; each reuse adapts a fresh
retrieval of these base states.

For clarity, the following token-level equations fix one crop step, decoder
layer, and KV head, and omit these indices. The operations are applied
independently at every layer and KV head; query-head indices are retained where
needed.

\subsection{Positional Key Rebinding (PKR)}
\label{sec:pkr}

Directly reused keys retain old positional encodings, creating a mismatch
with subsequent text queries (Section~\ref{sec:stale-binding}). \PKR{} corrects
this mismatch by shifting selected visual tokens to the current insertion
positions through a shared RoPE offset.

Let $\mathbf p_j^{\mathrm{src}}$ be the source RoPE position of selected
token $j$, and $\mathbf p_{\mathrm{ins}}$ the starting position of the new visual
KV block. Anchoring the first selected token there gives the common offset
$\boldsymbol\Delta=\mathbf p_{\mathrm{ins}}-\mathbf p_1^{\mathrm{src}}$
and target positions
$\mathbf p_j^{\mathrm{new}}=\mathbf p_j^{\mathrm{src}}+\boldsymbol\Delta$.
We apply RoPE to each retrieved clean key at its new position
(Figure~\ref{fig:framework}(b)):
\begin{equation}
  \mathbf k_j^{\mathrm{new}}
  =\mathbf R(\mathbf p_j^{\mathrm{new}})\bar{\mathbf k}_j,
  \label{eq:pkr}
\end{equation}
where $\mathbf k_j^{\mathrm{new}}$ is the rebound key and $\mathbf R(\cdot)$ is
the model's rotary position operator. For MRoPE, positions and offsets use the
model's native multidimensional coordinates \citep{bai2025qwen25vl}.

The common shift preserves relative positions among the selected visual tokens.
This invariance assumes a shared offset in the model's RoPE coordinates; it does
not hold in general for arbitrary token-wise reindexing.
By RoPE's rotation invariance, the shifted keys retain the norms and pairwise
inner products of the keys encoded at their source positions
\citep{su2021roformer}, while updating their positional relation to current text
queries. \PKR{} thus corrects the mismatch in Section~\ref{sec:stale-binding}
while preserving the computed key content. Each reuse starts from clean keys
and applies RoPE once at $\mathbf p_j^{\mathrm{new}}$, without recomputing content projections.

\subsection{Intent-guided Value Enhancement (IVE)}
\label{sec:ive}

After correcting positions, we also account for why the crop is requested.
Reasoning between observations typically identifies the next object, region, or
attribute to inspect. Full prefill lets new visual tokens interact with this
reasoning, whereas reuse retrieves values formed during an earlier observation
(Section~\ref{sec:prefill-gap}). \IVE{} uses the pre-crop reasoning's attention
to the requested region $B_t$ within its immediate parent observation $O_s$
to form a regional relevance prior and enhance the corresponding cached values
(Figure~\ref{fig:framework}(c)).

\paragraph{Reading visual attention from pre-crop reasoning.}
Let $\mathcal I$ be the reasoning-token window preceding the current crop
request, and $\mathcal H$ the query heads sharing the fixed KV head. For
$u\in\mathcal I$ and $h\in\mathcal H$, query $\mathbf q_{h,u}$ attends to
selected visual token $j$ within $B_t$ at parent-context index $x_j$.
For $x_j\in\mathcal C_u$, the attention weight is
\begin{equation}
  a_{h,u,j}
  =\frac{\exp\!\left(\mathbf q_{h,u}^{\top}\mathbf k_{x_j}/\sqrt d\right)}
  {\displaystyle\sum_{i\in\mathcal C_u}
    \exp\!\left(\mathbf q_{h,u}^{\top}\mathbf k_i/\sqrt d\right)}.
  \label{eq:attention}
\end{equation}
Here $a_{h,u,j}$ is the attention weight saved during decoding, $d$ is the
query/key head dimension, and $\mathcal C_u$ is the full causal context visible
to $u$. We set $a_{h,u,j}=0$ when $x_j\notin\mathcal C_u$.
The queries and keys already include RoPE. The index $x_j$ locates the
token in $O_s$'s visual block in the active context; it is distinct from the
source-cache index $\psi(j)$ and the token's RoPE coordinates. Full-context normalization retains
competition with other visual content and reasoning text.

\paragraph{Forming the regional relevance prior.}
For a nonempty reasoning window with positive attention mass on $B_t$, we
average over reasoning tokens and query heads sharing the fixed KV head, then
normalize across the $N\geq1$ selected visual tokens within $B_t$:
\begin{equation}
  b_j=\frac{\displaystyle\sum_{u\in\mathcal I}\sum_{h\in\mathcal H}a_{h,u,j}}
  {|\mathcal I|\,|\mathcal H|},
  \qquad
  \pi_j=\frac{b_j}{\displaystyle\sum_{i=1}^{N}b_i}.
  \label{eq:prior}
\end{equation}
Here $b_j$ is the aggregated attention and $\pi_j$ the token's relevance within
$B_t$. If the reasoning window is empty or regional attention
mass is zero or unavailable, we use $\pi_j=1/N$, yielding no enhancement. Aggregating
before regional normalization gives greater influence to reasoning text that
attends strongly to the target region, avoiding equal weight for text that
barely attends to it.

\paragraph{Enhancing values relevant to the current target.}
Using the uniform distribution $1/N$ as a baseline, we convert above-baseline
relevance into a gain $G_j$ and apply it to the retrieved base value:
\begin{equation}
  G_j=1+\beta\,\clip\!\left(N\pi_j-1,\,0,\,1\right),
  \qquad
  \mathbf v_j^{\mathrm{new}}=G_j\mathbf v_j^0.
  \label{eq:enhancement}
\end{equation}
Here $\clip(x,0,1)=\min(1,\max(0,x))$, and $\beta\geq0$ controls enhancement,
with $1\leq G_j\leq1+\beta$. Only tokens above the uniform relevance baseline
are enhanced; the others remain unchanged. Every reuse applies its gain to the
unchanged base value $\mathbf v_j^0$, avoiding cumulative amplification.
Finally, the keys $\mathbf k_j^{\mathrm{new}}$ and values
$\mathbf v_j^{\mathrm{new}}$ from Equations~\eqref{eq:pkr} and
\eqref{eq:enhancement} are inserted into the current cache, and decoding resumes.

\begin{table}[!t]
\captionsetup{font=small,skip=4pt}
\begin{minipage}{\textwidth}
  \centering
\caption{\textbf{Results on VisualProbe and HR-Bench 8K.}
Accuracy (\%; higher is better). Shaded rows compare Mini-o3, its ablation without intermediate crop images (\textit{w/o Interleaved images}), and our cache-reuse method. All shaded results report Avg@8, averaged over eight runs. Bold and underlining mark column-wise best and second-best results (including ties).}
  \label{tab:main_results}
  \label{tab:visualprobe_hr8k}
  \begin{adjustbox}{width=1.\textwidth}
  {
    \fontsize{7.5}{8.25}\selectfont
    \setlength{\tabcolsep}{1.5pt}
    \renewcommand{\arraystretch}{1.0}
    \setlength{\aboverulesep}{0.4ex}
    \setlength{\belowrulesep}{0.65ex}
    \input{tables/exp_visualprobe_hr8k}
  }
  \end{adjustbox}
  \renewcommand{\arraystretch}{1.0}
\end{minipage}

\vspace{1.5mm}

\begin{minipage}{\textwidth}
\caption{\textbf{Results on MME-RealWorld-Lite, V$^{*}$ Bench, and HR-Bench 4K.}
Accuracy (\%; higher is better). Shaded rows compare Mini-o3, its ablation without intermediate crop images (\textit{w/o Interleaved images}), and our cache-reuse method. All shaded results report Avg@8, averaged over eight runs. Bold and underlining mark column-wise best and second-best results (including ties).}
  \label{tab:additional_benchmarks}
  \centering
  \begin{adjustbox}{width=1.\textwidth}
  {
    \fontsize{7.5}{8.25}\selectfont
    \setlength{\tabcolsep}{1.5pt}
    \renewcommand{\arraystretch}{1.0}
    \setlength{\aboverulesep}{0.4ex}
    \setlength{\belowrulesep}{0.65ex}
    \input{tables/exp_hr_result}
  }
  \end{adjustbox}
  \renewcommand{\arraystretch}{1.0}
\end{minipage}
\par\vspace{6pt}
\begin{minipage}{\textwidth}
  \centering
  \includegraphics[width=\linewidth]{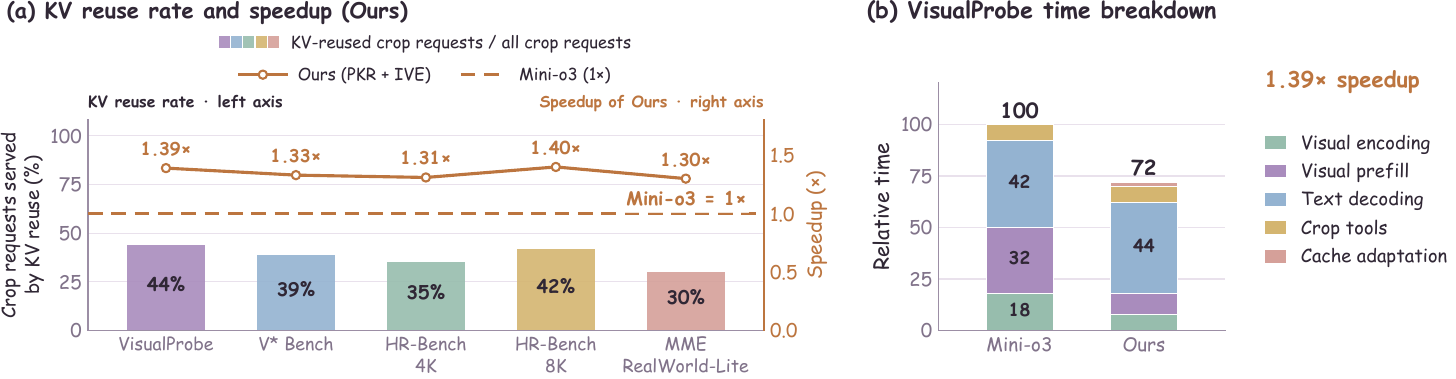}
  \captionsetup{type=figure,font=footnotesize,skip=3pt}
  \caption{\textbf{Inference efficiency of our method.} (a) KV reuse rate (\% of all crop requests) and end-to-end speedup of Ours over Mini-o3. (b) VisualProbe time breakdown, normalized to Mini-o3 = 100.}
  \label{fig:reuse_speedup}
\end{minipage}
\end{table}

\section{Experiments}
\label{sec:experiments}
\subsection{Experimental Setup}
\label{sec:experimental_setup}
\vspace{-3pt}

\paragraph{Benchmarks.}
We use VisualProbe~\citep{lai2026minio3} as our primary benchmark for
complex visual search, reporting accuracy on the Hard, Medium, and Easy
splits as well as overall accuracy. We also evaluate high-resolution visual
understanding on V$^*$ Bench~\citep{wu2024vstar} and HR-Bench
4K/8K~\citep{wang2025divide}, and general perception and reasoning on
MME-RealWorld-Lite~\citep{zhang2025mmerealworld}.

\paragraph{Baselines.}
We use Mini-o3~\citep{lai2026minio3} as the sole baseline for direct
comparison. Prior work reports nearly unchanged accuracy for DeepEyes,
Thyme, and Pixel Reasoner when tool-generated image outputs are suppressed
while retaining the preceding textual reasoning~\citep{yang2026position}.
In such settings, preserving accuracy while accelerating inference provides
limited evidence that KV reuse retains useful intermediate visual information. Our ablation shows
that removing cropped images causes a substantial performance drop
for Mini-o3. We therefore implement our method on Mini-o3 to evaluate
whether cache reuse reduces repeated re-encoding while preserving
multi-turn reasoning that depends on intermediate visual information.

\begin{figure}[t]
\centering
\begin{minipage}[t]{0.55\textwidth}
\vspace{0pt}
\centering
\includegraphics[width=\linewidth]{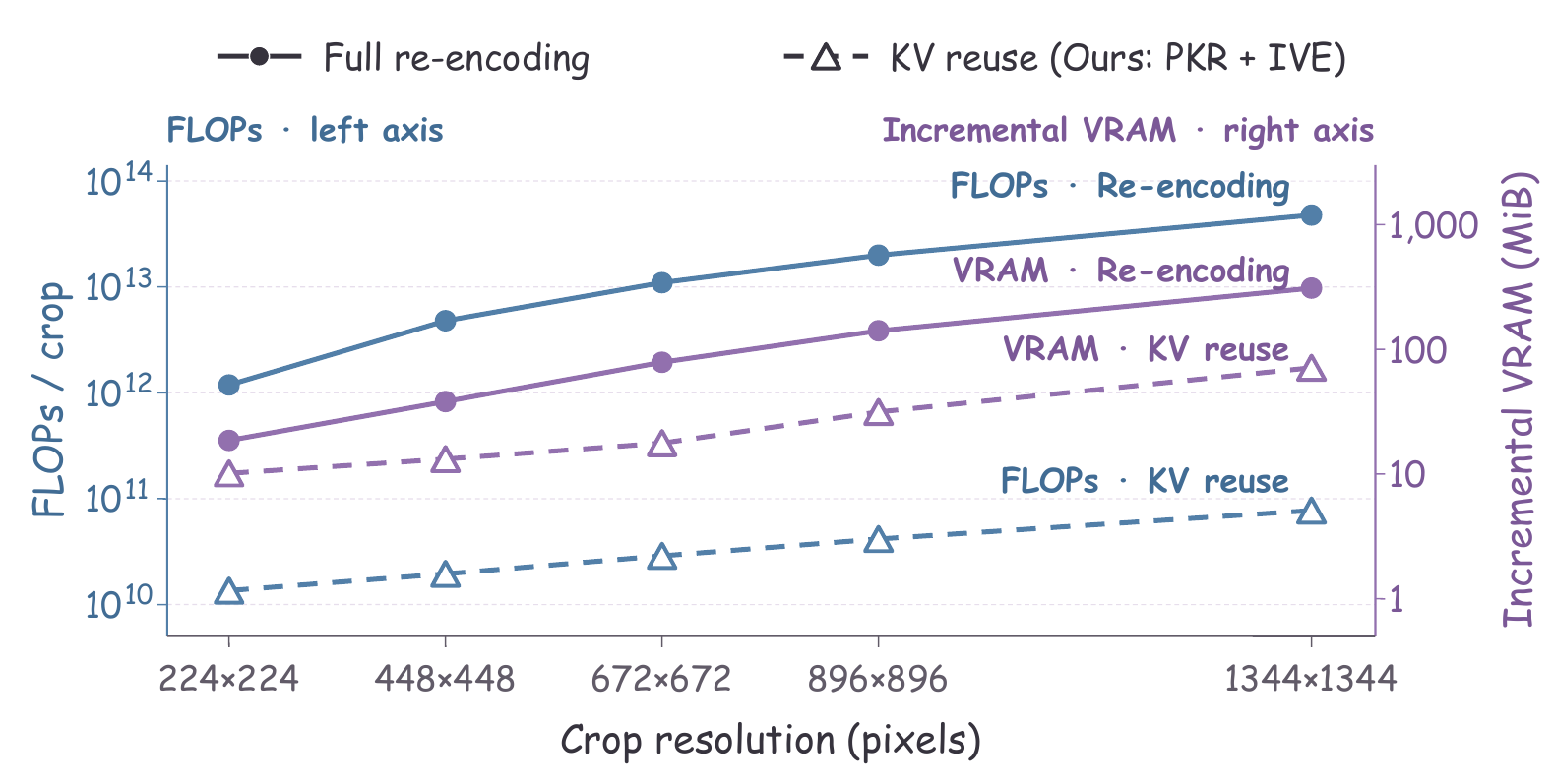}
\captionsetup{type=figure,font=footnotesize,position=bottom,skip=2pt}
\caption{\textbf{Crop processing costs on Mini-o3.} Full re-encoding versus PKR + IVE: FLOPs (left) and incremental VRAM (MiB, right), both on log scales.}
\label{fig:flops_resolution}
\end{minipage}\hfill
\begin{minipage}[t]{0.43\textwidth}
\vspace{0pt}
\centering
  \captionsetup{type=table,font=footnotesize,position=top,skip=2pt}
  \caption{\textbf{Incremental ablation} (accuracy, \%). Base: Mini-o3 without intermediate crop images; subsequent rows add PKR-based reuse and IVE. Bold marks column-wise best results.}
  \label{tab:incremental_ablation}
  \begingroup
  \setlength{\tabcolsep}{1.5pt}
  \renewcommand{\arraystretch}{2.25}
  \fontsize{7.2}{8.5}\selectfont
  \begin{tabular*}{\linewidth}{@{\extracolsep{\fill}}ccc|ccc@{}}
    \toprule
    \textbf{Base} & \textbf{PKR} & \textbf{IVE} &
    \makecell{VisualProbe\\Overall} &
    \makecell{V$^*$ Bench\\Overall} &
    \makecell{HR-Bench\\8K Overall} \\
    \midrule
    $\checkmark$ & & & 29.9 & 80.1 & 62.6 \\
    $\checkmark$ & $\checkmark$ & & 50.4 & 83.6 & 71.9 \\
    $\checkmark$ & $\checkmark$ & $\checkmark$ & \textbf{54.8} & \textbf{87.4} & \textbf{73.1} \\
    \bottomrule
  \end{tabular*}

  \endgroup

\end{minipage}
\end{figure}

\begin{figure}[t]
  \centering
  \includegraphics[width=\linewidth]{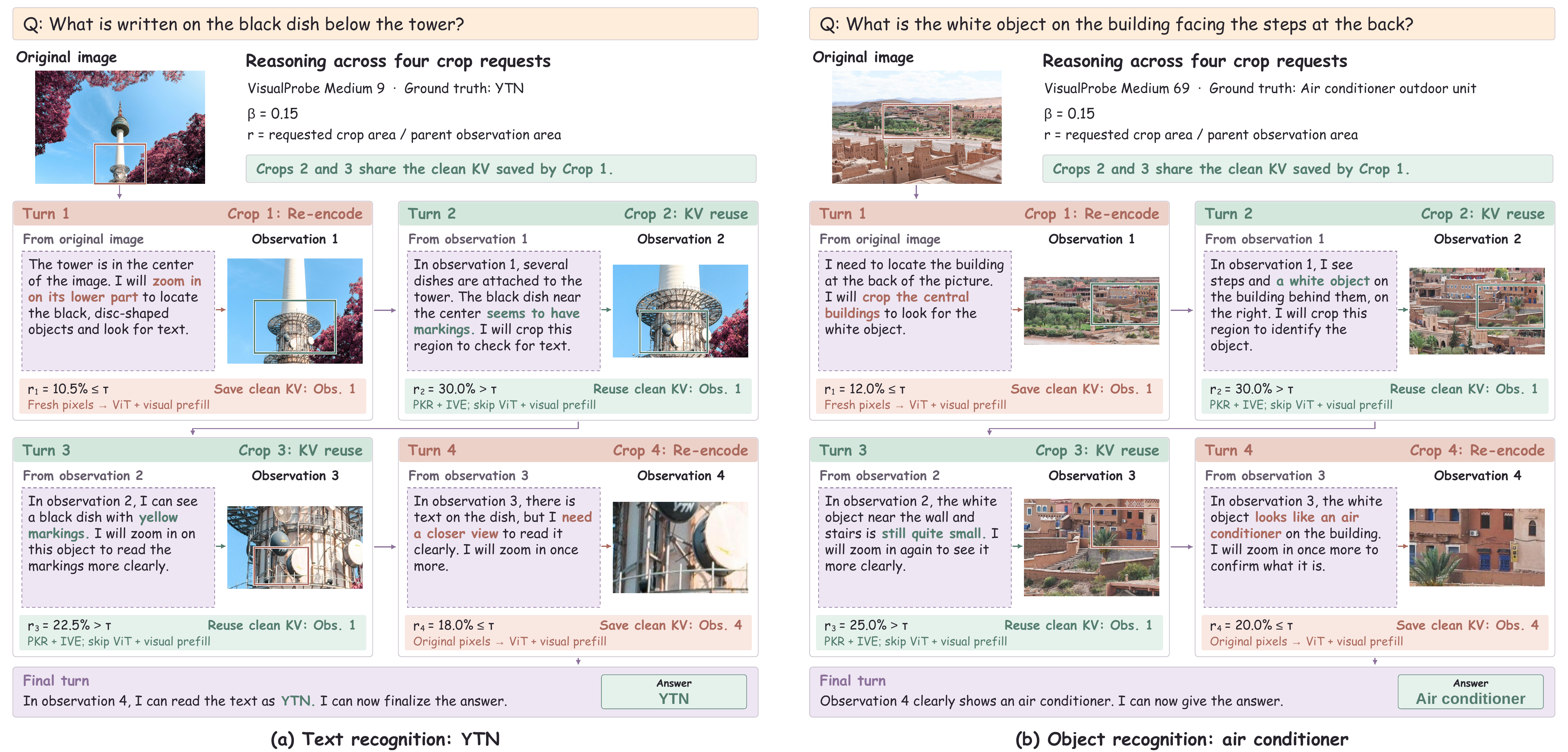}
  \caption{\textbf{Successive cropping with cache reuse.}
  (a) Reading text on a dish.
  (b) Identifying an air conditioner.
  Both examples use four crop requests, with KV reuse at
  the second and third steps.}
  \label{fig:successive_cropping}
\end{figure}

\paragraph{Implementation Details.}
We implement our method in Mini-o3's multi-turn vLLM~\citep{kwon2023vllm}
pipeline with frozen weights and no additional training. Routing uses the
crop-to-parent area ratio and cache availability, with $\tau=0.2$. We set
$\beta=0.15$ for IVE, bounding value gains within $[1,\,1.15]$.

\subsection{Main Results}
\label{sec:main_results}

Tables~\ref{tab:main_results} and~\ref{tab:additional_benchmarks} show that our method preserves or improves Mini-o3's overall accuracy across all five benchmarks, with gains of 0.2--1.3 percentage points. Removing intermediate crop images substantially degrades performance, including a 24.7-point drop on VisualProbe, confirming the importance of retaining intermediate visual evidence.

Figure~\ref{fig:reuse_speedup}(a) shows that serving 30--44\% of crop requests through KV reuse yields $1.30$--$1.40\times$ end-to-end speedups. Because crop requests differ in processing cost, the reuse ratio alone does not determine the end-to-end speedup. On VisualProbe, visual encoding and prefill together decrease from 50 to 18 normalized time units, while cache adaptation adds only 2 units. These savings outweigh the slightly longer text decoding, reducing total inference time by 28\% (Figure~\ref{fig:reuse_speedup}(b)). Figure~\ref{fig:flops_resolution} further shows that PKR + IVE reduces per-crop FLOPs and incremental VRAM relative to full re-encoding across all tested resolutions. Both paths become more costly as crop resolution increases, but the FLOPs gap widens in favor of reuse. Incremental VRAM also remains lower for reuse across the tested range, showing that the compute savings do not require higher per-crop memory use.

\subsection{Ablation Studies}
\label{sec:ablations}

To analyze the contributions of PKR and IVE, we start from Mini-o3 with intermediate crop images removed and progressively add the two components, as shown in Table~\ref{tab:incremental_ablation}. Enabling PKR-based visual KV reuse raises accuracy from 29.9\% to 50.4\% on VisualProbe, 80.1\% to 83.6\% on V$^*$ Bench, and 62.6\% to 71.9\% on HR-Bench 8K, recovering much of the performance lost without intermediate images. With routing unchanged, IVE adds 4.4, 3.8, and 1.2 percentage points, respectively, showing that current crop intent helps the model better use cached visual evidence.

\subsection{Qualitative Analysis}
\label{sec:qualitative_analysis}
\label{sec:qualitative}

\paragraph{Successive cropping with cache reuse.}
Figure~\ref{fig:successive_cropping} illustrates two examples
of text and object recognition.
In both cases, the first crop establishes the visual cache,
which supports the next two crop requests through PKR and IVE
without repeating visual encoding or visual-token prefill.
Although the third crop is requested from the second observation,
its visual states are retrieved from the cache established
by the first crop.
The fourth crop returns to the original pixels for full
re-encoding to obtain finer detail.
The resulting answers, ``YTN'' and ``air conditioner,''
match the ground truth, illustrating how reuse supports
intermediate inspection while re-encoding provides
additional detail for the final decision.

\par
\Needspace{24\baselineskip}
\begin{wrapfigure}{r}{0.6\textwidth}
    \centering
    \includegraphics[width=\linewidth]{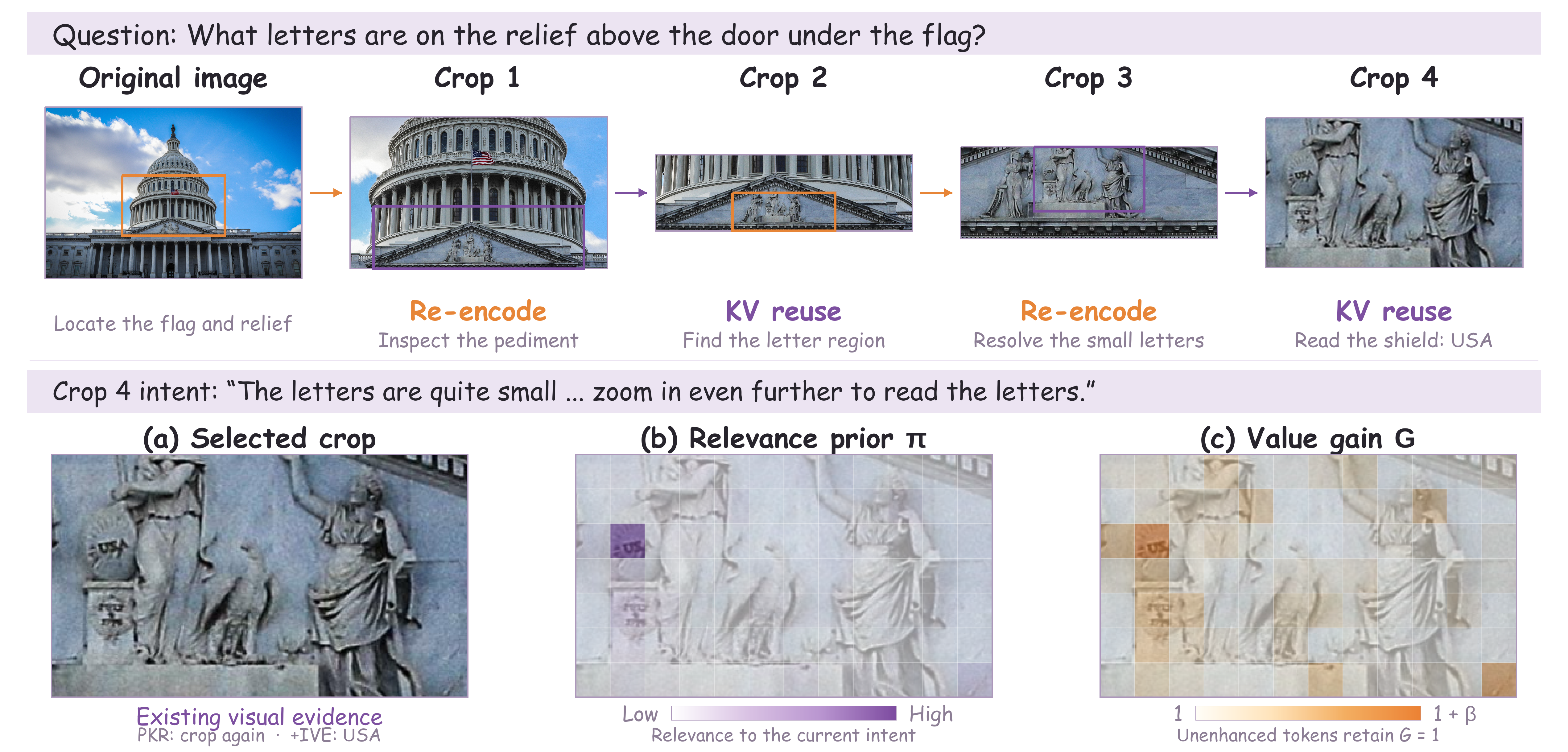}
    \caption{
    \textbf{Adaptive routing and intent-guided value enhancement.}
    Top: successive crops alternate between re-encoding and KV reuse.
    Bottom: (a) the selected fourth crop, (b) the intent-derived
    relevance prior $\pi$, and (c) the resulting value gains $G$.
    The model with PKR alone requests another crop; adding IVE yields the answer \texttt{USA}.
    }
    \label{fig:adaptive_ive_case}
\end{wrapfigure}

\paragraph{Intent-guided value enhancement.}
Figure~\ref{fig:adaptive_ive_case} illustrates a multi-turn search for the letters on a relief beneath a flag, where successive crops narrow the view from the building to the target region. Our framework re-encodes Crops~1 and~3 to obtain finer visual details and reuses cached KV states for Crops~2 and~4 to avoid repeated computation. The fourth crop further highlights the role of IVE: although the cached visual evidence already contains the letters \texttt{USA}, the model with PKR alone still requests another crop; with IVE, it instead answers \texttt{USA} using the available evidence. The visualizations show that the pre-crop reasoning intent to zoom in and read the letters produces a relevance prior that emphasizes the letter-bearing region. IVE converts this prior into bounded gains on the corresponding cached values, strengthening their contributions during subsequent decoding. This example illustrates that the presence of relevant visual information in the cache does not guarantee its effective use: adapting reused values to the current reasoning intent can help the model extract an answer from existing evidence.

\section{Conclusion}
\label{sec:conclusion}

We present an adaptive visual KV cache reuse framework for efficient
multi-turn multimodal reasoning. The framework selects between regional
cache reuse and full re-encoding, reducing repeated visual processing
while preserving access to additional image detail. Our analysis connects
two challenges arising from skipped visual prefill: stale positional
bindings in cached keys and the absence of intervening crop intent in
historical values. Positional Key Rebinding corrects the former by
assigning cached keys their current positions, while Intent-guided Value
Enhancement addresses the latter by modulating value contributions using
pre-crop reasoning attention. These adaptations require no additional
training and leave the model's crop-decision mechanism unchanged. On reuse hits, the
framework resumes reasoning from adapted visual states without repeating
visual encoding or decoder prefill for the selected visual tokens.

\subsection*{AI use statement}

We used generative AI tools to assist with Chinese-to-English
translation, manuscript drafting and revision, refining the
exposition of the method and its mathematical analysis,
literature search, reference formatting, and figure layout
refinement. The authors reviewed the AI-assisted content
and verified the technical statements and references.
The authors take full responsibility for the final manuscript,
including all claims, results, and supporting materials.

\par

\bibliographystyle{plainnat}
\bibliography{references}

\end{document}

%% file: sections/observations_motivation.tex
% !TEX root = ../main.tex
\section{Observations and Motivation}
\label{sec:observations}

\begin{figure}[t]
  \centering
  \includegraphics[width=0.9\linewidth]{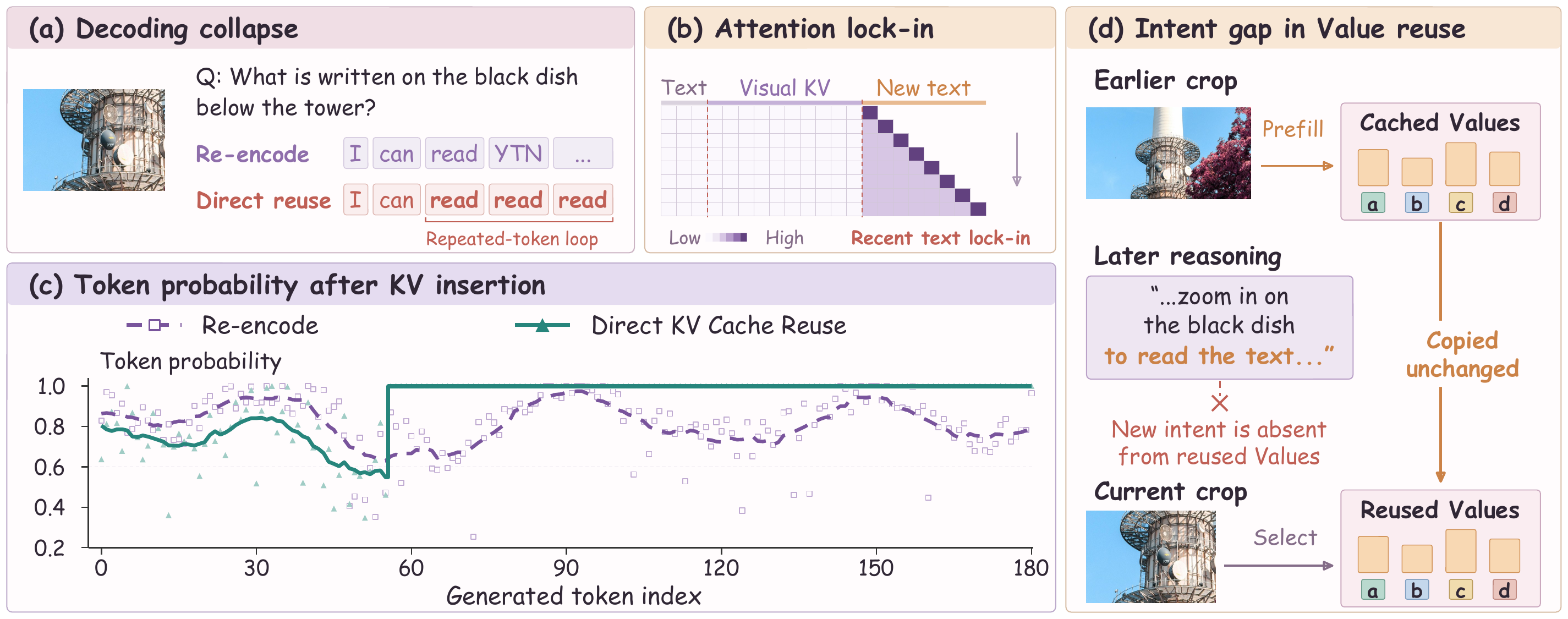}
  \caption{\textbf{Why visual KV reuse requires adaptation.}
  (a,b) Outputs and attention patterns illustrate repetitive decoding
  and attention concentrated on recent text under direct reuse.
  (c) Generated-token probabilities abruptly approach one under direct
  reuse but remain variable after full re-encoding (Re-encode).
  (d) Skipping visual prefill leaves reused values without the intervening
  crop intent.}
  \label{fig:motivation}
      \vspace{-1.2em}
\end{figure}

Successive crops often cover overlapping image regions, creating an opportunity
to reuse visual KV states and avoid repeated computation. However, these states
were generated in an earlier reasoning context, and whether they can be directly
reused for a later crop remains unclear. We therefore begin by examining the
behavior of direct reuse and skipped visual prefill in Figure~\ref{fig:motivation}
to identify the conditions needed for effective visual cache reuse.

\subsection{Stale Positional Bindings and Decoding Collapse}
\label{sec:stale-binding}

Direct insertion of historical visual KV can lead to repetitive token loops
(Figure~\ref{fig:motivation}(a,b)). To understand this behavior, we examine how the position encoded in a cached key
affects attention scores. Omitting layer and head indices, let $\mathbf p_j^{\mathrm{src}}$ be
the source position of selected visual token $j$ and $\bar{\mathbf k}_j$ its key
before rotation. The historical cache stores
$\mathbf k_j^{\mathrm{src}}=\mathbf R(\mathbf p_j^{\mathrm{src}})\bar{\mathbf k}_j$.
For a subsequent text token at position $\mathbf p_u$, with pre-rotation query
$\bar{\mathbf q}_u$, direct reuse gives
\begin{equation}
s_{u,j}^{\mathrm{direct}}
=\frac{\bigl(\mathbf R(\mathbf p_u)\bar{\mathbf q}_u\bigr)^\top
\mathbf R(\mathbf p_j^{\mathrm{src}})\bar{\mathbf k}_j}{\sqrt d}
=\frac{\bar{\mathbf q}_u^\top
\mathbf R(\mathbf p_j^{\mathrm{src}}-\mathbf p_u)\bar{\mathbf k}_j}{\sqrt d}.
\label{eq:stale-score}
\end{equation}
Here $d$ is the head dimension and $\mathbf R(\cdot)$ is the RoPE rotation operator.
Equation~\eqref{eq:stale-score} shows that relocating a cache entry does not change
the position encoded within its key. At the current insertion position
$\mathbf p_j^{\mathrm{new}}$, the query--key interaction requires the relative rotation
$\mathbf R(\mathbf p_j^{\mathrm{new}}-\mathbf p_u)$, whereas direct reuse still uses
$\mathbf R(\mathbf p_j^{\mathrm{src}}-\mathbf p_u)$. This mismatch can alter attention matching
even when the content projections remain fixed.

The resulting shift also affects the readout from the entire context. Let
$\mathcal C_u$ denote the full context visible to the current token, including
historical text and the newly inserted visual KV states. Attention weights and
their output are (with $s_{u,i}=\mathbf q_u^\top\mathbf k_i/\sqrt d$,
where $\mathbf q_u$ and $\mathbf k_i$ include RoPE, $\mathbf v_i$ is the value
at context index $i$, and $\mathbf o_u$ is the head output)
\begin{equation}
\begin{aligned}
\alpha_{u,i}
&=\frac{\exp(s_{u,i})}{\displaystyle\sum_{v\in\mathcal C_u}\exp(s_{u,v})},
\qquad
\mathbf o_u=\sum_{i\in\mathcal C_u}\alpha_{u,i}\mathbf v_i.
\end{aligned}
\label{eq:context-attention}
\end{equation}
All attention weights in Equation~\eqref{eq:context-attention} share a softmax
denominator, so changing reused visual-key scores also redistributes attention
to reasoning text. This affects next-token predictions and may accumulate as
generated tokens enter the context, reinforcing recent outputs.
Figure~\ref{fig:motivation}(a,b) shows token repetition and attention lock-in;
Figure~\ref{fig:motivation}(c) contrasts probabilities approaching one under direct reuse with continued
variation after full re-encoding. These effects motivate correcting positions
while retaining cached key content. Section~\ref{sec:pkr} introduces Positional
Key Rebinding (\PKR), which uses RoPE's rotation properties to encode clean
cached keys at their current insertion positions.

\subsection{The Semantic Gap Left by Skipping Visual Prefill}
\label{sec:prefill-gap}

Position correction aligns old keys with new positions but does not restore
prefill's semantic interaction. After inspecting a tower, for example, the model
may request a crop of the black dish below it to read its text
(Figure~\ref{fig:motivation}(d)). Full prefill places the crop's visual tokens
after this reasoning, allowing the new objective to influence later-layer
representations through causal attention. Direct value reuse skips this interaction.

Let $(\mathbf E_{\mathrm{src}},\mathcal P_{\mathrm{src}})$ denote the source
observation's visual input and preceding context, and
$(\mathbf E_{\mathrm{crop}},\mathcal P_{\mathrm{crop}})$ those of the current
crop. Only $\mathcal P_{\mathrm{crop}}$ includes the intervening reasoning.
Let $\psi(j)$ map crop token $j$ to its source-cache index. Omitting layer and
head indices, the value projection $\mathbf W_V$ acts on pre-projection visual
states $\mathbf h$, giving
\begin{equation}
\begin{aligned}
\mathbf v_j^{\mathrm{prefill}}
&=\mathbf W_V\mathbf h_j(\mathbf E_{\mathrm{crop}};\mathcal P_{\mathrm{crop}}),\\
\mathbf v_j^{\mathrm{reuse}}
&=\mathbf v_j^0
=\mathbf W_V\mathbf h_{\psi(j)}(\mathbf E_{\mathrm{src}};\mathcal P_{\mathrm{src}}).
\end{aligned}
\label{eq:prefill-values}
\end{equation}
Here $\mathbf v_j^{\mathrm{prefill}}$ is newly computed for crop token $j$,
$\mathbf v_j^{\mathrm{reuse}}$ is its directly reused value, and $\mathbf v_j^0$
is the unchanged value retrieved from the source cache. These cached values
predate the new crop intent and may originate from an older fully prefilled
ancestor. Current queries can adjust attention weights but still read these
earlier representations. Section~\ref{sec:ive} introduces Intent-guided Value
Enhancement (\IVE): pre-crop reasoning attention strengthens cached values
relevant to the current objective without visual prefill.

%% file: tables/exp_visualprobe_hr8k.tex
% Equal metric columns; all scores are pending re-evaluation.
\newcommand{\resulthead}[1]{\makebox[\linewidth][c]{#1}}
\begin{tabular}{L{.285}|*{4}{C{.10214286}}|*{3}{C{.10214286}}}
\toprule
\multirow{2}{*}{\textbf{Model}}
& \multicolumn{4}{c|}{\textbf{VisualProbe}}
& \multicolumn{3}{c}{HR-Bench 8K} \\
& \resulthead{Hard} & \resulthead{Medium} & \resulthead{Easy} & \resulthead{\textbf{Overall}}
& \resulthead{\textbf{FSP}} & \resulthead{\textbf{FCP}} & \resulthead{\textbf{Overall}} \\
\midrule
LLaVA-OneVision~\cite{avkrtable_li2024llavaonevision} & 13.4 & 12.5 & 36.2 & 19.2 & 67.3 & 52.3 & 59.8 \\
Qwen2.5-VL~\cite{bai2025qwen25vl} & 23.9 & 26.0 & 39.1 & 29.2 & 78.8 & 51.8 & 65.3 \\
Pixel-Reasoner~\cite{wang2025pixelreasoner} & 28.8 & 29.6 & 58.4 & 37.3 & 80.0 & 54.3 & 66.9 \\
Thyme~\cite{zhang2026thyme} & 46.2 & 43.3 & 62.4 & 49.1 & 86.5 & \underline{57.5} & 72.0 \\
DeepEyes~\cite{zheng2026deepeyes} & 35.1 & 29.8 & 60.1 & 39.2 & 86.8 & \textbf{58.5} & \underline{72.6} \\
\midrule
\ours{Mini-o3~\cite{lai2026minio3}} & \ours{\underline{49.1}} & \ours{\textbf{49.6}} & \ours{\underline{68.1}} & \ours{\underline{54.6}} & \ours{\underline{87.5}} & \ours{56.1} & \ours{71.8} \\
\ours{\hspace{1em}w/o Interleaved images} & \ours{\textit{18.9}} & \ours{\textit{23.5}} & \ours{\textit{50.4}} & \ours{\textit{29.9}} & \ours{\textit{70.3}} & \ours{\textit{55.0}} & \ours{\textit{62.6}} \\
\ours{\textbf{Mini-o3 + PKR + IVE (Ours)}} & \ours{\textbf{50.0}} & \ours{\underline{48.5}} & \ours{\textbf{70.2}} & \ours{\textbf{54.8}} & \ours{\textbf{87.7}} & \ours{\textbf{58.5}} & \ours{\textbf{73.1}} \\
\bottomrule
\end{tabular}

%% file: tables/exp_hr_result.tex
% Equal metric columns; all scores are pending re-evaluation.
\newcommand{\resulthead}[1]{\makebox[\linewidth][c]{\fontsize{7}{8.25}\selectfont #1}}
\begin{tabular}{L{.285}|*{3}{C{.07944444}}|*{3}{C{.07944444}}|*{3}{C{.07944444}}}
\toprule
\multirow{2}{*}{\textbf{Model}}
& \multicolumn{3}{c|}{\textbf{MME-RealWorld-Lite}}
& \multicolumn{3}{c|}{$V^*$ Bench}
& \multicolumn{3}{c}{HR-Bench 4K} \\
& \resulthead{Perception} & \resulthead{Reasoning} & \resulthead{\textbf{Overall}}
& \resulthead{\textbf{Attr}} & \resulthead{\textbf{Spatial}} & \resulthead{\textbf{Overall}}
& \resulthead{\textbf{FSP}} & \resulthead{\textbf{FCP}} & \resulthead{\textbf{Overall}} \\
\midrule
LLaVA-OneVision~\cite{avkrtable_li2024llavaonevision} & 52.8 & 42.7 & 48.5 & 75.7 & 75.0 & 75.4 & 72.0 & 54.0 & 63.0 \\
Qwen2.5-VL~\cite{bai2025qwen25vl} & 48.8 & 37.7 & 44.1 & 73.9 & 67.1 & 71.2 & 85.2 & 52.2 & 68.8 \\
Pixel-Reasoner~\cite{wang2025pixelreasoner} & 53.1 & 44.5 & 49.7 & 83.5 & 76.3 & 80.6 & 86.0 & 60.3 & 72.9 \\
Thyme~\cite{zhang2026thyme} & 59.1 & 49.1 & 55.2 & 83.5 & 80.3 & 82.2 & 91.0 & 63.0 & \underline{77.0} \\
DeepEyes~\cite{zheng2026deepeyes} & 56.6 & 50.5 & 54.3 & \textbf{91.3} & \textbf{88.2} & \textbf{90.1} & \underline{91.3} & 59.0 & 75.1 \\
\midrule
\ours{Mini-o3~\cite{lai2026minio3}} & \ours{\underline{66.8}} & \ours{\underline{59.9}} & \ours{\underline{64.1}} & \ours{\underline{89.4}} & \ours{82.9} & \ours{86.8} & \ours{89.5} & \ours{\textbf{63.7}} & \ours{76.6} \\
\ours{\hspace{1em}w/o Interleaved images} & \ours{\textit{53.4}} & \ours{\textit{46.5}} & \ours{\textit{50.7}} & \ours{\textit{82.6}} & \ours{\textit{76.3}} & \ours{\textit{80.1}} & \ours{\textit{81.5}} & \ours{\textit{\underline{63.3}}} & \ours{\textit{72.4}} \\
\ours{\textbf{Mini-o3 + PKR + IVE (Ours)}} & \ours{\textbf{67.2}} & \ours{\textbf{60.3}} & \ours{\textbf{64.5}} & \ours{87.8} & \ours{\underline{86.8}} & \ours{\underline{87.4}} & \ours{\textbf{91.7}} & \ours{62.7} & \ours{\textbf{77.2}} \\
\bottomrule
\end{tabular}

%% file: references.bib
@inproceedings{zheng2026deepeyes,
  title = {{DeepEyes}: Incentivizing ``Thinking with Images'' via Reinforcement Learning},
  author = {Zheng, Ziwei and Yang, Minghao and Hong, Jack and Zhao, Chenxiao and Xu, Guohai and Yang, Le and Shen, Chao and Yu, Xing},
  booktitle = {International Conference on Learning Representations},
  year = {2026},

}

@inproceedings{lai2026minio3,
  title = {{Mini-o3}: Scaling Up Reasoning Patterns and Interaction Turns for Visual Search},
  author = {Lai, Xin and Li, Junyi and Li, Wei and Liu, Tao and Li, Tianjian and Zhao, Hengshuang},
  booktitle = {International Conference on Learning Representations},
  year = {2026},

}

@article{yao2024cacheblend,
  title = {{CacheBlend}: Fast Large Language Model Serving for {RAG} with Cached Knowledge Fusion},
  author = {Yao, Jiayi and Li, Hanchen and Liu, Yuhan and Ray, Siddhant and Cheng, Yihua and Zhang, Qizheng and Du, Kuntai and Lu, Shan and Jiang, Junchen},
  journal = {arXiv preprint arXiv:2405.16444},
  year = {2024},
  eprint = {2405.16444},
  archivePrefix = {arXiv},
  primaryClass = {cs.LG},
  doi = {10.48550/arXiv.2405.16444}
}

@article{zhao2025mpic,
  title = {{MPIC}: Position-Independent Multimodal Context Caching System for Efficient {MLLM} Serving},
  author = {Zhao, Shiju and Hu, Junhao and Huang, Rongxiao and Zheng, Jiaqi and Chen, Guihai},
  journal = {arXiv preprint arXiv:2502.01960},
  year = {2025},
  eprint = {2502.01960},
  archivePrefix = {arXiv},
  primaryClass = {cs.LG},
  doi = {10.48550/arXiv.2502.01960}
}

@article{ma2026kamera,
  title = {{Kamera}: Unified Position-Invariant Multimodal {KV} Cache for Training-Free Reuse},
  author = {Ma, Bole and Eitzinger, Jan and Koestler, Harald and Wellein, Gerhard},
  journal = {arXiv preprint arXiv:2606.23581},
  year = {2026},
  eprint = {2606.23581},
  archivePrefix = {arXiv},
  primaryClass = {cs.DC},
  doi = {10.48550/arXiv.2606.23581}
}

@article{chen2026conduit,
  title = {{CONDUIT}: A Unified Residual-Stream Restoration Framework for {KV} Cache Reuse in Vision-Language Models},
  author = {Chen, Pengan and Zheng, Kaisheng and Hong, Liang and Yi, Lixia and Jiang, Jiyue and Chen, Jiayang and Wang, Yixuan and Fan, Yimin and Liu, Xinyuan and Li, Jiayi and Zhang, Zhanqiu and Guo, Yiwen and Li, Yu},
  journal = {arXiv preprint arXiv:2609.05821},
  year = {2026},
  eprint = {2609.05821},
  archivePrefix = {arXiv},
  primaryClass = {cs.CL},
  doi = {10.48550/arXiv.2609.05821}
}

@article{su2021roformer,
  title = {{RoFormer}: Enhanced Transformer with Rotary Position Embedding},
  author = {Su, Jianlin and Lu, Yu and Pan, Shengfeng and Murtadha, Ahmed and Wen, Bo and Liu, Yunfeng},
  journal = {arXiv preprint arXiv:2104.09864},
  year = {2021},
  eprint = {2104.09864},
  archivePrefix = {arXiv},
  primaryClass = {cs.CL},
  doi = {10.48550/arXiv.2104.09864}
}

@article{bai2025qwen25vl,
  title = {{Qwen2.5-VL} Technical Report},
  author = {Bai, Shuai and others},
  journal = {arXiv preprint arXiv:2502.13923},
  year = {2025},
  eprint = {2502.13923},
  archivePrefix = {arXiv},
  primaryClass = {cs.CV},
  doi = {10.48550/arXiv.2502.13923}
}

@inproceedings{wu2024vstar,
  title = {{V*}: Guided Visual Search as a Core Mechanism in Multimodal {LLMs}},
  author = {Wu, Penghao and Xie, Saining},
  booktitle = {Proceedings of the IEEE/CVF Conference on Computer Vision and Pattern Recognition},
  year = {2024}
}

@inproceedings{hu2024sketchpad,
  title = {Visual {Sketchpad}: Sketching as a Visual Chain of Thought for Multimodal Language Models},
  author = {Hu, Yushi and Shi, Weijia and Fu, Xingyu and Roth, Dan and Ostendorf, Mari and Zettlemoyer, Luke and Smith, Noah A. and Krishna, Ranjay},
  booktitle = {Advances in Neural Information Processing Systems},
  volume = {37},
  year = {2024}
}

@article{wang2025pixelreasoner,
  title = {Pixel Reasoner: Incentivizing Pixel-Space Reasoning with Curiosity-Driven Reinforcement Learning},
  author = {Wang, Haozhe and Su, Alex and Ren, Weiming and Lin, Fangzhen and Chen, Wenhu},
  journal = {arXiv preprint arXiv:2505.15966},
  year = {2025}
}

@inproceedings{gao2025icot,
  title = {Interleaved-Modal Chain-of-Thought},
  author = {Gao, Jun and Li, Yongqi and Cao, Ziqiang and Li, Wenjie},
  booktitle = {Proceedings of the IEEE/CVF Conference on Computer Vision and Pattern Recognition},
  pages = {19520--19529},
  year = {2025}
}

@inproceedings{chen2024fastv,
  title = {An Image is Worth 1/2 Tokens After Layer 2: Plug-and-Play Inference Acceleration for Large Vision-Language Models},
  author = {Chen, Liang and Zhao, Haozhe and Liu, Tianyu and Bai, Shuai and Lin, Junyang and Zhou, Chang and Chang, Baobao},
  booktitle = {European Conference on Computer Vision},
  year = {2024}
}

@inproceedings{zhang2025sparsevlm,
  title = {{SparseVLM}: Visual Token Sparsification for Efficient Vision-Language Model Inference},
  author = {Zhang, Yuan and Fan, Chun-Kai and Ma, Junpeng and Zheng, Wenzhao and Huang, Tao and Cheng, Kuan and Gudovskiy, Denis A. and Okuno, Tomoyuki and Nakata, Yohei and Keutzer, Kurt and Zhang, Shanghang},
  booktitle = {Proceedings of the 42nd International Conference on Machine Learning},
  series = {Proceedings of Machine Learning Research},
  volume = {267},
  pages = {74840--74857},
  year = {2025}
}

@inproceedings{li2024snapkv,
  title = {{SnapKV}: {LLM} Knows What You are Looking for Before Generation},
  author = {Li, Yuhong and Huang, Yingbing and Yang, Bowen and Venkitesh, Bharat and Locatelli, Acyr and Ye, Hanchen and Cai, Tianle and Lewis, Patrick and Chen, Deming},
  booktitle = {Advances in Neural Information Processing Systems},
  volume = {37},
  year = {2024}
}

@article{ye2026betadartspp,
  title = {{\ensuremath{\beta}-DARTS++}: Bi-Level Regularization for Proxy-Robust Differentiable Architecture Search},
  author = {Ye, Peng and He, Tong and Li, Baopu and Chen, Tao and Bai, Lei and Ouyang, Wanli},
  journal = {IEEE Transactions on Pattern Analysis and Machine Intelligence},
  year = {2026},
  volume = {48},
  number = {2},
  pages = {1407--1423},
  month = feb,
  doi = {10.1109/TPAMI.2025.3616249}
}

@article{ye2025stimulativepp,
  title = {{Stimulative Training++}: Go Beyond the Performance Limits of Residual Networks},
  author = {Ye, Peng and He, Tong and Tang, Shengji and Li, Baopu and Chen, Tao and Bai, Lei and Ouyang, Wanli},
  journal = {IEEE Transactions on Pattern Analysis and Machine Intelligence},
  year = {2025},
  volume = {47},
  number = {8},
  pages = {6463--6479},
  month = aug,
  doi = {10.1109/TPAMI.2025.3558421}
}

@article{ye2025mow,
  title = {Dynamic Model Merging With Mixture of Weights},
  author = {Ye, Peng and Huang, Chenyu and Shen, Mingzhu and Chen, Tao and Huang, Yongqi and Ouyang, Wanli},
  journal = {IEEE Transactions on Circuits and Systems for Video Technology},
  year = {2025},
  volume = {35},
  number = {8},
  pages = {7925--7939},
  month = aug,
  doi = {10.1109/TCSVT.2025.3544680}
}

@inproceedings{zhuang2025vasparse,
  title = {{VASparse}: Towards Efficient Visual Hallucination Mitigation via Visual-Aware Token Sparsification},
  author = {Zhuang, Xianwei and Zhu, Zhihong and Xie, Yuxin and Liang, Liming and Zou, Yuexian},
  booktitle = {Proceedings of the IEEE/CVF Conference on Computer Vision and Pattern Recognition},
  pages = {4189--4199},
  year = {2025}
}

@inproceedings{yang2025visionzip,
  title = {{VisionZip}: Longer is Better but Not Necessary in Vision Language Models},
  author = {Yang, Senqiao and Chen, Yukang and Tian, Zhuotao and Wang, Chengyao and Li, Jingyao and Yu, Bei and Jia, Jiaya},
  booktitle = {Proceedings of the IEEE/CVF Conference on Computer Vision and Pattern Recognition},
  pages = {19792--19802},
  year = {2025}
}

@inproceedings{zhang2023h2o,
  title = {{H$_2$O}: Heavy-Hitter Oracle for Efficient Generative Inference of Large Language Models},
  author = {Zhang, Zhenyu and Sheng, Ying and Zhou, Tianyi and Chen, Tianlong and Zheng, Lianmin and Cai, Ruisi and Song, Zhao and Tian, Yuandong and R{\'e}, Christopher and Barrett, Clark and Wang, Zhangyang and Chen, Beidi},
  booktitle = {Advances in Neural Information Processing Systems},
  volume = {36},
  year = {2023},
  doi = {10.52202/075280-1506}
}

@inproceedings{xiao2024streamingllm,
  title = {Efficient Streaming Language Models with Attention Sinks},
  author = {Xiao, Guangxuan and Tian, Yuandong and Chen, Beidi and Han, Song and Lewis, Mike},
  booktitle = {The Twelfth International Conference on Learning Representations},
  year = {2024}
}

@inproceedings{shang2025prumerge,
  title = {{LLaVA-PruMerge}: Adaptive Token Reduction for Efficient Large Multimodal Models},
  author = {Shang, Yuzhang and Cai, Mu and Xu, Bingxin and Lee, Yong Jae and Yan, Yan},
  booktitle = {Proceedings of the IEEE/CVF International Conference on Computer Vision},
  pages = {22857--22867},
  year = {2025}
}

@article{avkrtable_li2024llavaonevision,
  author = {Li, Bo and Zhang, Yuanhan and Guo, Dong and Zhang, Renrui and Li, Feng and Zhang, Hao and Zhang, Kaichen and Zhang, Peiyuan and Li, Yanwei and Liu, Ziwei and Li, Chunyuan},
  title = {{LLaVA-OneVision}: Easy Visual Task Transfer},
  journal = {arXiv preprint arXiv:2408.03326},
  year = {2024},
  doi = {10.48550/arXiv.2408.03326},

}

@inproceedings{yang2026position,
  title = {Position: Your {VLM} May Not Be Thinking with Interleaved Images},
  author = {Yang, Wenjie and Zhu, Siqi and Huang, Zengfeng},
  booktitle = {Forty-third International Conference on Machine Learning Position Paper Track},
  year = {2026},

}

@inproceedings{kwon2023vllm,
  title = {Efficient Memory Management for Large Language Model Serving with {PagedAttention}},
  author = {Kwon, Woosuk and Li, Zhuohan and Zhuang, Siyuan and Sheng, Ying and Zheng, Lianmin and Yu, Cody Hao and Gonzalez, Joseph E. and Zhang, Hao and Stoica, Ion},
  booktitle = {Proceedings of the 29th Symposium on Operating Systems Principles},
  year = {2023},
  pages = {611--626},
  doi = {10.1145/3600006.3613165}
}

@inproceedings{wang2025divide,
  title = {Divide, Conquer and Combine: A Training-Free Framework for High-Resolution Image Perception in Multimodal Large Language Models},
  author = {Wang, Wenbin and Ding, Liang and Zeng, Minyan and Zhou, Xiabin and Shen, Li and Luo, Yong and Yu, Wei and Tao, Dacheng},
  booktitle = {Proceedings of the AAAI Conference on Artificial Intelligence},
  volume = {39},
  number = {8},
  pages = {7907--7915},
  year = {2025}
}

@inproceedings{zhang2025mmerealworld,
  title = {{MME-RealWorld}: Could Your Multimodal {LLM} Challenge High-Resolution Real-World Scenarios that are Difficult for Humans?},
  author = {Zhang, Yi-Fan and Zhang, Huanyu and Tian, Haochen and Fu, Chaoyou and Zhang, Shuangqing and Wu, Junfei and Li, Feng and Wang, Kun and Wen, Qingsong and Zhang, Zhang and Wang, Liang and Jin, Rong and Tan, Tieniu},
  booktitle = {International Conference on Learning Representations},
  year = {2025},

}

@inproceedings{zhang2026thyme,
  title = {{Thyme}: Think Beyond Images},
  author = {Zhang, Yi-Fan and Lu, Xingyu and Yin, Shukang and Fu, Chaoyou and Chen, Wei and Hu, Xiao and Wen, Bin and Jiang, Kaiyu and Liu, Changyi and Zhang, Tianke and Fan, Haonan and Chen, Kaibing and Chen, Jiankang and Ding, Haojie and Tang, Kaiyu and Zhang, Zhang and Wang, Liang and Yang, Fan and Gao, Tingting and Zhou, Guorui},
  booktitle = {International Conference on Learning Representations},
  year = {2026},

}
